\documentclass[runningheads]{llncs}
\usepackage{graphicx}

\usepackage{tikz}
\usepackage{comment}
\usepackage{amsmath,amssymb} 
\usepackage{color}

\usepackage[accsupp]{axessibility}  
\usepackage{epsfig}
\usepackage{amsmath,amsfonts} 
\usepackage{amssymb}
\usepackage{booktabs}
\usepackage{dsfont}
\usepackage{enumitem} 
\usepackage{array} 
\usepackage{url} 
\usepackage[font=scriptsize]{subfig}
\usepackage{pifont}
\usepackage{mmstyle}
\usepackage{breakcites}

\newcommand{\cmark}{\ding{51}}%
\newlength\savewidth\newcommand\shline{\noalign{\global\savewidth\arrayrulewidth\global\arrayrulewidth 1pt}\hline\noalign{\global\arrayrulewidth\savewidth}}

\usepackage[pagebackref,breaklinks,colorlinks]{hyperref}

\usepackage[capitalize]{cleveref}
\crefname{section}{Sec.}{Secs.}
\Crefname{section}{Section}{Sections}
\Crefname{table}{Table}{Tables}
\crefname{table}{Tab.}{Tabs.}

\begin{document}
\pagestyle{headings}
\mainmatter
\def\ECCVSubNumber{5542}  

\title{Point Scene Understanding via Disentangled Instance Mesh Reconstruction}

\titlerunning{Point Scene Understanding via DIMR}
%
\author{
Jiaxiang Tang\inst{1} \and
Xiaokang Chen\inst{1} \and
Jingbo Wang\inst{2} \and
Gang Zeng\inst{1}
}
\authorrunning{J. Tang et al.}
%

\institute{
Key Laboratory of Perception (MoE), School of AI, Peking University \\
\and 
Chinese University of Hong Kong \\
\email{\{tjx, pkucxk\}@pku.edu.cn,~ wj020@ie.cuhk.edu.hk,~ zeng@pku.edu.cn}
}

\maketitle

\begin{abstract}

Semantic scene reconstruction from point cloud is an essential and challenging task for 3D scene understanding. 
This task requires not only to recognize each instance in the scene, but also to recover their geometries based on the partial observed point cloud. 
Existing methods usually attempt to directly predict occupancy values of the complete object based on incomplete point cloud proposals from a detection-based backbone.
However, this framework always fails to reconstruct high fidelity mesh due to the obstruction of various detected false positive object proposals and the ambiguity of incomplete point observations for learning occupancy values of complete objects. 
To circumvent the hurdle, we propose a Disentangled Instance Mesh Reconstruction (DIMR) framework for effective point scene understanding.
A segmentation-based backbone is applied to reduce false positive object proposals, which further benefits our exploration on the relationship between recognition and reconstruction. 
Based on the accurate proposals, we leverage a mesh-aware latent code space to disentangle the processes of shape completion and mesh generation, relieving the ambiguity caused by the incomplete point observations. 
Furthermore, with access to the CAD model pool at test time, our model can also be used to improve the reconstruction quality by performing mesh retrieval without extra training.
We thoroughly evaluate the reconstructed mesh quality with multiple metrics, and demonstrate the superiority of our method on the challenging ScanNet dataset.
Code is available at \url{https://github.com/ashawkey/dimr}.

\end{abstract}

\section{Introduction}

Semantic scene reconstruction can facilitate numerous real-world applications, such as robot navigation, AR/VR and interior design. This task aims to understand the semantic information of each object and recover their geometries from partial observations (\eg point cloud from 3D scans). Several previous methods only focus on object recognition in the scene~\cite{jiang2020pointgroup,he2021dyco3d,chen2021hierarchical,hu2021bidirectional,hu2021vmnet,choy20194d,wang2017cnn} by semantic and instance segmentation, or the completion of the partial observed point cloud~\cite{senushkin2020decoder,huang2019indoor,han2019deep,dai2018scancomplete,dai2021spsg,dai2020sg}. In order to further explore both semantic and geometry information, in this paper, we aim to jointly complete these two different tasks in one framework.

\begin{figure}[t!]
    \centering
    \includegraphics[width=\textwidth]{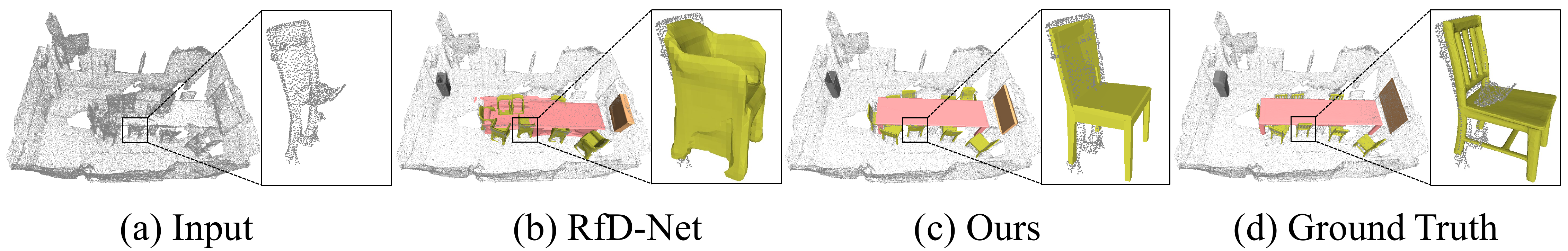}
    \caption{{Point Scene Instance Mesh Reconstruction}. 
    With an incomplete point cloud scene as input, our method learns to recognize each object instance and reconstruct a complete mesh that matches the input observation well. 
    }
    \label{fig:teaser}
\end{figure}

Recently, researchers begin to explore the relationship of semantic and geometry information for scene understanding. 
Semantic Scene Completion~\cite{song2017semantic,liu2018satnet,guo2018vvnet,chen20203d} reconstructs occluded geometry by performing semantic segmentation for both visible and occluded space in dense voxel grids. 
Similarly, RevealNet~\cite{hou2020revealnet} performs instance segmentation in dense voxel grids. 
Due to the demanding memory requirement, these works are typically limited by the low-resolution dense voxel grids and can not reconstruct high fidelity objects in the scene. 
RfD-Net~\cite{Nie_2021_CVPR} first proposes to work directly on sparse point clouds, which can recognize and reconstruct objects in high-resolution mesh representation. 
However, as shown in Figure~\ref{fig:teaser}(b), this Reconstruction-from-Detection pipeline always fails to reconstruct high fidelity objects. 
In general, the reason can be mainly categorized into two aspects. 
The first one is the numerous false positive proposals from the detection module. 
These false positive proposals cause the mismatch between the incomplete point clouds and the complete mesh, thus obstructing training an effective shape completion network. 
The second one is the structure ambiguity caused by incomplete point observations for directly learning occupancy values of complete objects. 
Therefore, to further explore this problem, we should answer the following two questions for this task: 
\emph{i}) Does the accurate foreground object proposals improve the reconstruction quality? 
\emph{ii}) How to mitigate the structure ambiguity of incomplete point cloud for mesh reconstruction?

\begin{figure*}[t!]
    \centering
    \includegraphics[width=\textwidth]{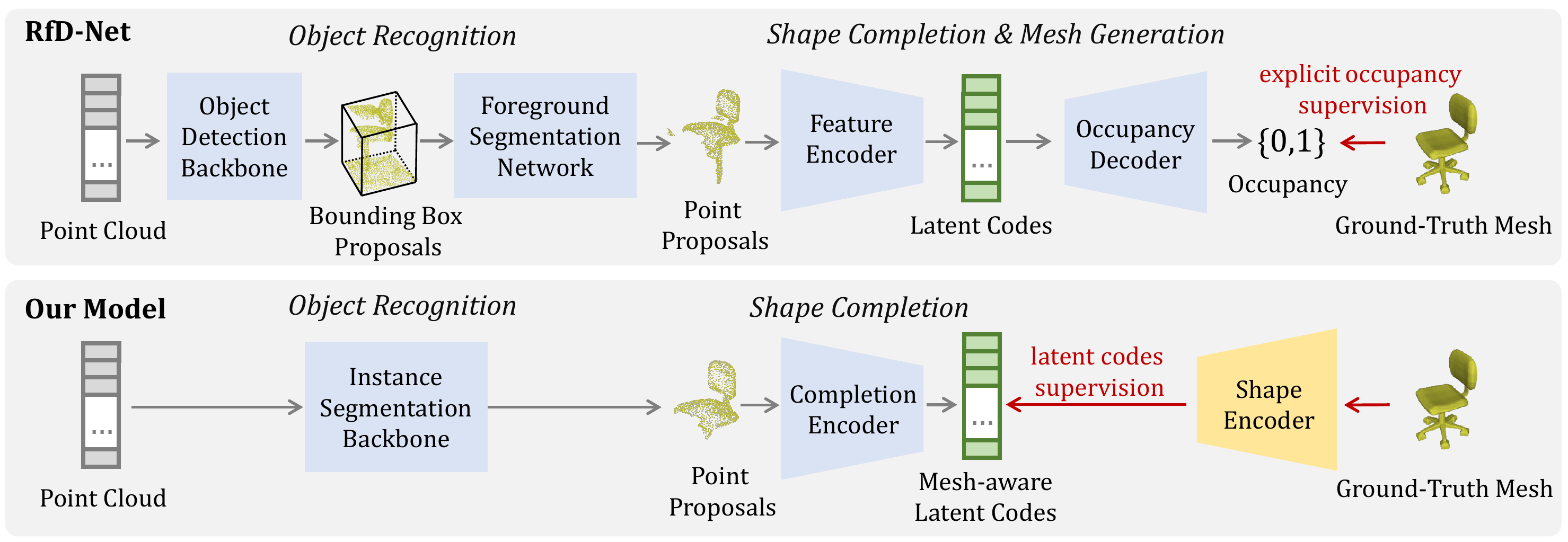}
    \caption{{A comparison of the training process between our method and RfD-Net}. 
    For object recognition, RfD-Net uses a detection-based backbone to predict bounding boxes, then extracts the foreground points with a segmentation network, while we apply a straight-forward instance segmentation backbone.
    Furthermore, we disentangle shape completion and mesh generation by adopting a pre-trained mesh autoencoder and supervise shape completion by latent codes. The mesh generation part is only used during inference (not shown in the figure), mitigating the issue of difficult explicit occupancy supervision in RfD-Net. 
    Without harming the end-to-end training of object recognition and shape completion, our method significantly improves the quality of generated meshes.
     }
    \label{fig:comp}
\end{figure*}

In this paper, we propose our Disentangled Instance Mesh Reconstruction framework to answer these two questions. 
Our pipeline contains two stages, namely instance segmentation and instance mesh reconstruction. 
Comparing against state-of-the-art point cloud detection approaches~\cite{qi2019deep,xie2020mlcvnet,liu2021group}, we observe that instance segmentation framework~\cite{jiang2020pointgroup,chen2021hierarchical} can reduce the false positive rate of object proposal significantly. 
Therefore, we generate the object proposal for object completion based on instance segmentation.
With the proposals from the instance segmentation framework, the quality of completed objects is improved consequentially. 
For the second question, we propose a disentangled instance mesh reconstruction approach to recover high fidelity mesh of each incomplete object. 
Different from directly learning occupancy values of complete objects based on incomplete point observations~\cite{Nie_2021_CVPR}, we propose to disentangle the shape completion and mesh generation for mesh reconstruction. 
The shape completion module aims to recover the necessary structure information of incomplete objects to mitigate the ambiguity for high-fidelity mesh reconstruction, which is the goal of the mesh generation module. 
Especially, our shape completion module does not focus on direct completion of mesh, due to the noisy information from input point observations. 
More effectively, we propose the mesh-aware latent code as the supervision for our shape completion module. 
The target latent code is encoded from the complete point cloud by a pre-trained encoder, and can be used for mesh reconstruction by a pre-trained decoder, namely the mesh generator. 
Therefore, the structure information of the complete point cloud for mesh reconstruction is encoded into our mesh-aware latent code, and our shape completion module can learn this structure information directly. 
After training, the structure information of incomplete point observations can be recovered by this module and mitigate the ambiguity for mesh reconstruction. 
With the pre-trained mesh generator, our method can generate high-quality meshes consistent with the point observations as shown in Figure~\ref{fig:teaser} (c).
Furthermore, if we have access to the CAD model pool, our model can be used to search the nearest neighbors in the latent code space to perform mesh retrieval or assist mesh generation, without the need of extra training.

To summarize, the contributions of this paper are as follows:
\begin{itemize}
    \item We analyze the weaknesses of the previous detection based framework and propose a new pipeline for point scene instance mesh reconstruction, which first performs instance segmentation on incomplete point scenes and then completes each object instance with a mesh that matches the observed points.
    
    \item We design a disentangled instance mesh reconstruction strategy to mitigate the ambiguity of learning complete shapes from incomplete point cloud observations, by leveraging a mesh-aware latent code space. 
    Furthermore, it can also be used for mesh retrieval with the access to a provided model pool.
    
    \item We studied multiple metrics to measure the performance in mesh completion quality, and proposed a new metric to measure point-to-mesh mapping quality. 
    Results show that our method performs better than previous state-of-the-arts on the challenging ScanNet dataset, especially on complex structures such as chairs and tables.
\end{itemize}

\section{Related Work}

\subsection{Point Scene Instance Segmentation}
Instance segmentation has been an important topic for point scene understanding with the availability of large-scale point cloud scene datasets.
Current methods can be categorized into detection-based and segmentation-based methods.
Detection-based methods~\cite{engelmann20203d,hou20193d,yang2019learning} first regress 3D bounding boxes and then mask out background points inside each box to get the final instance segmentation. 
However, the two-step pipeline is not straightforward and usually inefficient.
Instead, segmentation-based methods~\cite{jiang2020pointgroup,wang2018sgpn,pham2019jsis3d,he2021dyco3d,lahoud20193d,han2020occuseg,chen2021hierarchical,liang2021instance,zhong2022hiera} directly predict semantic segmentation and then cluster points into instance proposals.
For example, PointGroup~\cite{jiang2020pointgroup} uses sparse 3D CNNs~\cite{SubmanifoldSparseConvNet,yan2018second,liu2019point,tang2020searching} to extract point cloud features, and propose a dual-set clustering algorithm to better distinguish the void space between object instances.
Later works~\cite{chen2021hierarchical,he2021dyco3d} mainly focuses on more efficient and concise instance clustering algorithms such as dynamic convolution and hierarchical aggregation.
We choose the segmentation-based backbone for its simplicity, and bridge instance segmentation to mesh reconstruction in end-to-end training. 

\subsection{3D Shape Completion}

\noindent \textbf{Object Completion.}
This line of research mainly focuses on shape completion of single objects.
Many works~\cite{wang2020cascaded,tchapmi2019topnet,nie2020skeleton,yuan2018pcn} focus on the completion of point cloud shapes, with incomplete point clouds as the input and completed point clouds as the output.
However, these methods usually complete up to a limited number of points which is not enough to represent high resolution shapes due to the sampling problem.
Other works choose dense voxel grids~\cite{stutz2018learning,dai2017shape,han2017high} or implicit functions~\cite{chibane2020implicit,chen2019learning} to perform shape completion.
Many works~\cite{chen2019learning,mo2019structurenet,li2017grass,wu2020pq,paschalidou2021neural} adopt an autoencoder architecture to learn a compact latent code for each shape.
BSP-Net~\cite{chen2020bspnet,bspnet_tpami} proposes to approximate shapes with a Binary Space Partitioning (BSP) tree, which shows good results on mesh reconstruction from dense voxel grids and single view images.

\noindent \textbf{Scene Completion.}
Instead of focusing on single objects, scene completion aims to complete all objects from a partial observation such as a 3D scan.
Early works usually start from volumetric representations and the completion task can be viewed as a dense labeling task on voxel grids.
Semantic Scene Completion~\cite{song2017semantic,guo2018vvnet,liu2018satnet,chen20203d,dai2018scancomplete,chen2020real,tang2021not} voxelizes the point cloud into dense voxel grids and predicts semantic labels of all voxels in both visible and occluded regions. 
RevealNet~\cite{hou2020revealnet} proposes semantic instance completion, which performs object detection on voxel grids and then completes each instance within the cropped voxel grids.
Other works focus on mesh representations.
Total 3D understanding~\cite{nie2020total3dunderstanding,zhang2021holistic,engelmann2021points} performs object detection and mesh reconstruction on RGB images.
RfD-Net~\cite{Nie_2021_CVPR} first performs semantic instance completion on point clouds directly and generates completed instance meshes. 
It adopts a detection-based backbone and uses implicit functions for mesh reconstruction, demonstrating that these two tasks are complementary.
Assuming the availability of a CAD model pool, CAD retrieval aims to find the best-fitting CAD models and align them to the point scenes~\cite{avetisyan2019scan2cad,avetisyan2020scenecad,dahnert2019joint,hampali2021monte,grabner20183d}, images~\cite{kuo2020mask2cad}, or videos~\cite{maninis2020vid2cad}, optionally allowing for deformation of single objects~\cite{uy2021joint,ishimtsev2020cad}.
Our method follows this line of research, with point clouds as the direct input and reconstructed instance meshes as the output. 
Differently, we separate shape completion and mesh generation tasks to ease the training process with the proposed latent instance mesh reconstruction.

\section{Method}
\begin{figure*}[ht!]
    \centering
    \includegraphics[width=\textwidth]{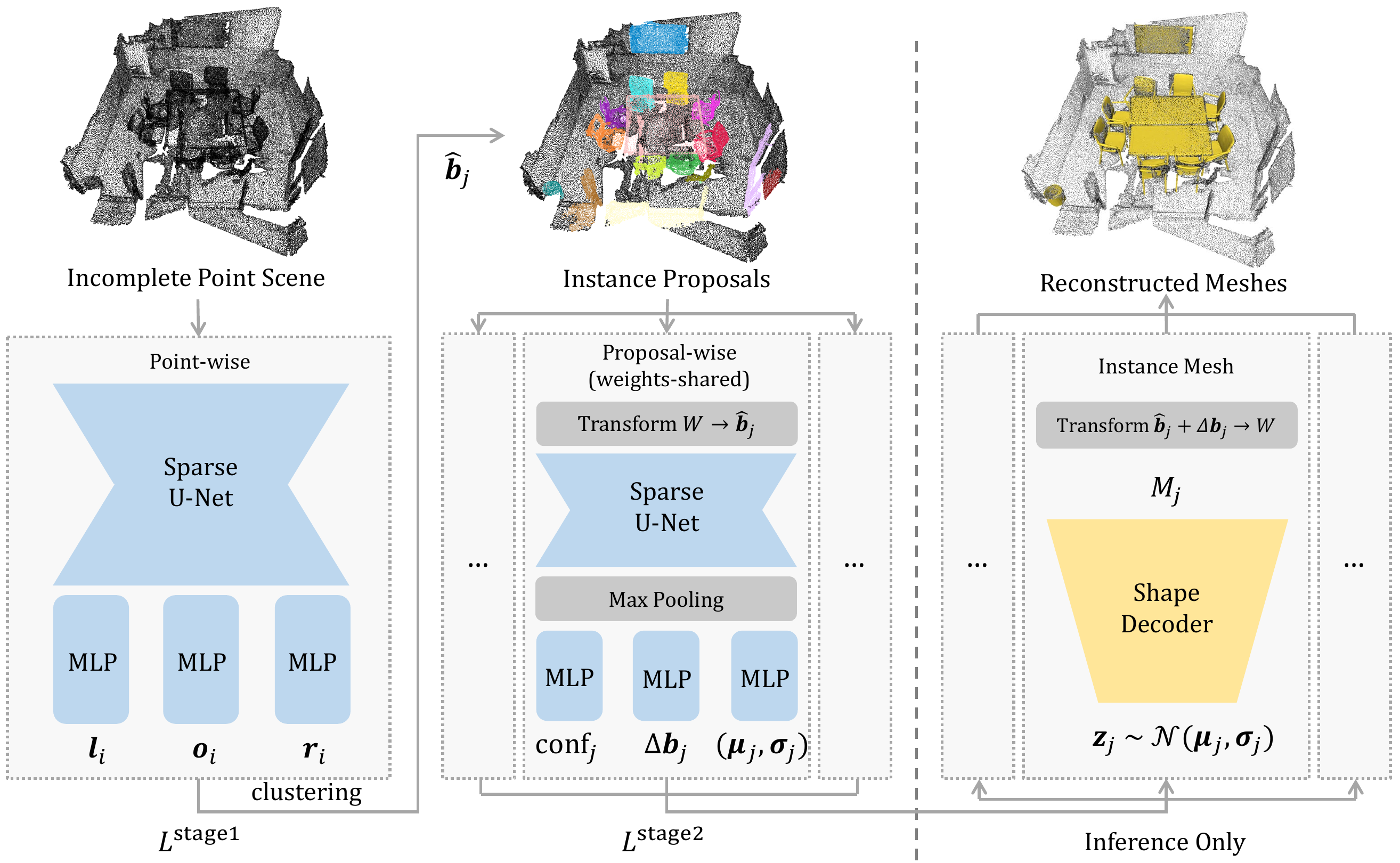}
    \caption{{Overview of the network architecture}. 
    The network first learns point-wise features including semantic labels $\textbf{l}_i$, instance offsets $\textbf{o}_i$ and instance rotations $\textbf{r}_i$.
    Then, instance proposals are clustered and fed to the second stage for learning proposal-wise features including confidence scores $\text{conf}_j$, residual bounding boxes $\Delta \textbf{b}_j$, and latent shape distributions $\mathcal N (\boldsymbol \mu_j, \boldsymbol \sigma_j)$.
    During inference, instance meshes are generated by decoding the mesh-aware latent codes sampled from the latent shape distributions.
    }
    \label{fig:arch}
\end{figure*}

We introduce our pipeline as illustrated in Figure~\ref{fig:arch}.
Overall, the pipeline consists of two training stages: point-wise learning and proposal-wise learning.
The first stage (Section~\ref{sec:pointwise}) takes a sparse 3D CNN backbone to perform point-wise predictions including semantic labels, instance center offsets, and rotation angles.
In the second stage (Section~\ref{sec:proposalwise}), another sparse 3D CNN is used to predict proposal-wise results including residual bounding boxes, confidence scores, and the latent distributions of complete meshes.
Only in inference, the latent codes sampled from these distributions are decoded to generate compact meshes (Section~\ref{sec:meshgen}), which are transformed back to the world coordinate system with the refined bounding boxes to compose the final reconstructed scene.

\subsection{Learning Point-wise Features}
\label{sec:pointwise}
In this stage, we focus on learning point-wise features including the semantic labels for semantic segmentation, offsets from the instance center for instance segmentation, and instance rotation angles for bounding box regression.
We follow~\cite{jiang2020pointgroup} and use a sparse 3D U-Net~\cite{SubmanifoldSparseConvNet,UNet} as the backbone to extract features. 
The input to the network is a point set $\mathbb P = \{\mathbf{p}_1, \mathbf{p}_2, \cdots, \mathbf{p}_N\}$, where each point is described by its coordinate $\mathbf{p}_i = (x_i, y_i, z_i), i \in [1, N]$.
These points are voxelized before being fed to the backbone. 
To obtain the per-point feature, we map the voxel feature from the backbone back to the point and get $\mathbf{F}_{\text{point}} \in \mathbb R^{N \times {D_\text{point}}}$, where $D_\text{point}$ is the feature dimension.
Then, three Multi-Layer Perceptrons (MLPs) are applied to regress three point-wise targets respectively.
For the semantic label, the prediction is the classification logits $\mathbf l_i \in \mathbb R^C$, where $C$ is the total number of classes.
For the instance offset, the prediction is the offset ${\mathbf o_i = (\Delta x_i, \Delta y_i, \Delta z_i)}$ from the current point to the instance center it belongs to.
In addition to these two regular heads for instance segmentation~\cite{jiang2020pointgroup,chen2021hierarchical}, we use a third head to predict the orientation of the instance that covers the current point to build an approximate oriented bounding box.
We only predict the rotation angle $r_i \in [-\pi, \pi)$ along the $z$-axis following~\cite{Nie_2021_CVPR}, since the rotation along $x,y$-axes for most instances can be ignored.

To optimize the afore-mentioned objectives, we use the cross-entropy loss $L_\text{cls}^\text{semantic}$ for semantic segmentation, the L1 Loss $L_\text{reg}^\text{offset}$ for instance offset regression, and follow~\cite{Nie_2021_CVPR,huang2018cooperative} to disentangle the angle loss into a hybrid of classification and regression loss $\mathcal L_\text{cls}^\text{angle} + \mathcal L_\text{reg}^\text{angle}$. 
So far, the loss function for the first stage $\mathcal L^\text{stage1}$ can be concluded as the sum of these four parts.

\subsection{Learning Proposal-wise Features}
\label{sec:proposalwise}
The second stage handles proposal-wise predictions that bridge instance segmentation to instance mesh reconstruction.
Given the point-wise predictions from stage one, we first apply a clustering algorithm~\cite{jiang2020pointgroup} to group the whole scene's point cloud into $L$ instance point cloud proposals $\mathcal P = \{P_1, P_2, \cdots, P_L\}$.
We then transform each instance point cloud to its canonical coordinate system for better proposal-wise feature learning~\cite{shi2019pointrcnn}.
Specifically, each instance point cloud $P_j \in \mathcal P$ is:
1) recentered at the mean instance center $\mathbf{\bar c}_j = \frac 1 {|P_j|} \sum_{i \in P_j} (\mathbf{p}_i + \mathbf{o}_i)$;
2) rotated along $z$-axis for the negative mean rotation angle $-\bar r_j = -\frac 1 {|P_j|} \sum_{i \in P_j} r_i$ to make the instance front-facing; and
3) scaled into $[0, 1]$ on each axis by dividing $\mathbf{s}_j$ , where $\mathbf{s}_j \in \mathbb R^3$ is the approximate instance scale calculated from the minimum and maximum coordinates of the rotated points.
Another voxelization is applied on each instance proposal to extract proposal-wise features.
The transformed instance points with their features $\mathbf{F}_{\text{point}}$ are fed into the second sparse 3D U-Net, after which a max-pooling layer is used to output the proposal-wise features $\mathbf{F}_{\text{prop}} \in \mathbb R^{L \times D_\text{prop}}$, where $D_\text{prop}$ is the feature dimension. 
This allows the point-wise features learned in stage one to be smoothly propagated to later modules.
The original scale information $\mathbf{s}_j$ is preserved by being concatenated to the features.

To reconstruct instance meshes from $\mathbf{F}_{\text{prop}}$, we still need to regress three targets: proposal confidence, residual bounding box and latent shape distributions.

\noindent \textbf{Proposal confidence.}
An MLP followed by a sigmoid function is applied to regress the confidence value $\text{conf}_j \in [0, 1]$ for proposal $P_j$. 
The ground truth for the confidence is decided by the largest point Intersection over Union (IoU) between the proposal and ground-truth instances following~\cite{jiang2020pointgroup}.

\noindent \textbf{Residual bounding Box.}
From the first stage, we already have an initial 7 Degree-of-Freedom (DoF) oriented bounding box $\mathbf{\hat b}_j = \{ \bar r_j, \mathbf{\bar c}_j, \mathbf{s}_j \}$.
However, this bounding box is inaccurate due to partial observation and occlusion, especially for the instance scale (\textit{e.g.}, missing of chair legs leads to underestimated scale on the $z$-axis).
We therefore use an MLP to predict a residual bounding box $\Delta \mathbf{b}_j$ to refine this initial bounding box, 
and the final bounding box is given by $\mathbf{b}_j = \mathbf{\hat b}_j + \Delta \mathbf{b}_j$.

\noindent \textbf{Latent shape distributions.}
To address the ambiguity problem in shape completion, a probabilistic generative model is usually adopted~\cite{wu2016learning,achlioptas2018learning,Nie_2021_CVPR}.
We take a similar way by assuming that the complete shape is sampled from a latent Gaussian distribution, and learn it through the reparameterization trick~\cite{kingma2013auto}.
An MLP is used to regress the mean and standard deviation $\boldsymbol \mu_j, \boldsymbol \sigma_j \in \mathbb R^{D_\text{shape}}$, where $D_\text{shape}$ is the latent shape code dimension.
To allow supervised learning, we need to know the ground-truth latent distribution $(\boldsymbol \mu^\text{gt}_j, \boldsymbol \sigma^\text{gt}_j)$ of ground-truth meshes, which will be described in Section~\ref{sec:meshgen}.

\begin{figure}[t!]
    \centering
    
    \subfloat[Disentangled Instance Mesh Reconstruction]{
        \includegraphics[width=0.48\linewidth]{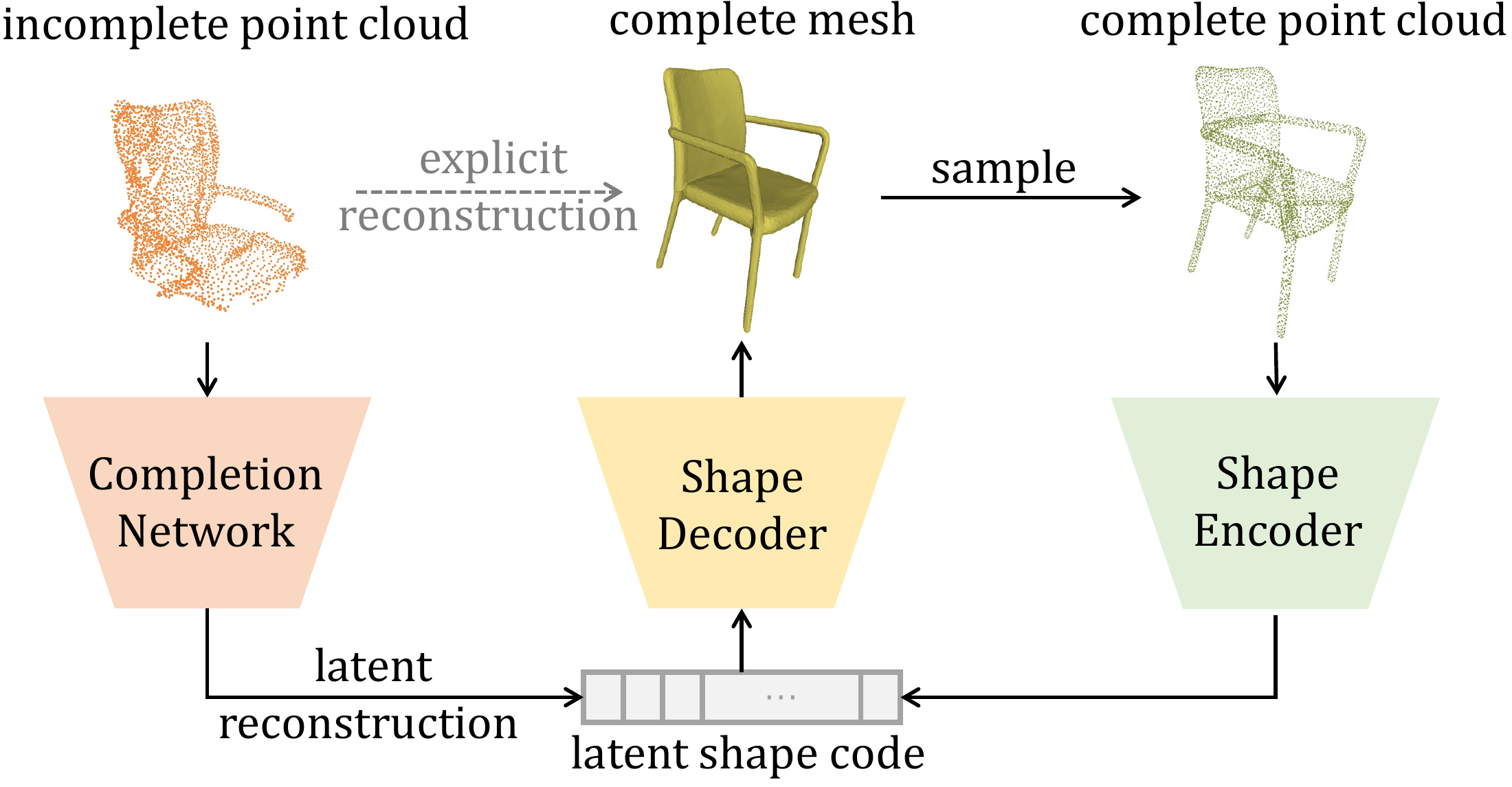}
        \label{fig:iimr}
    }
    \hfill
    \subfloat[Mesh Retrieval and Generation]{
        \includegraphics[width=0.455\linewidth]{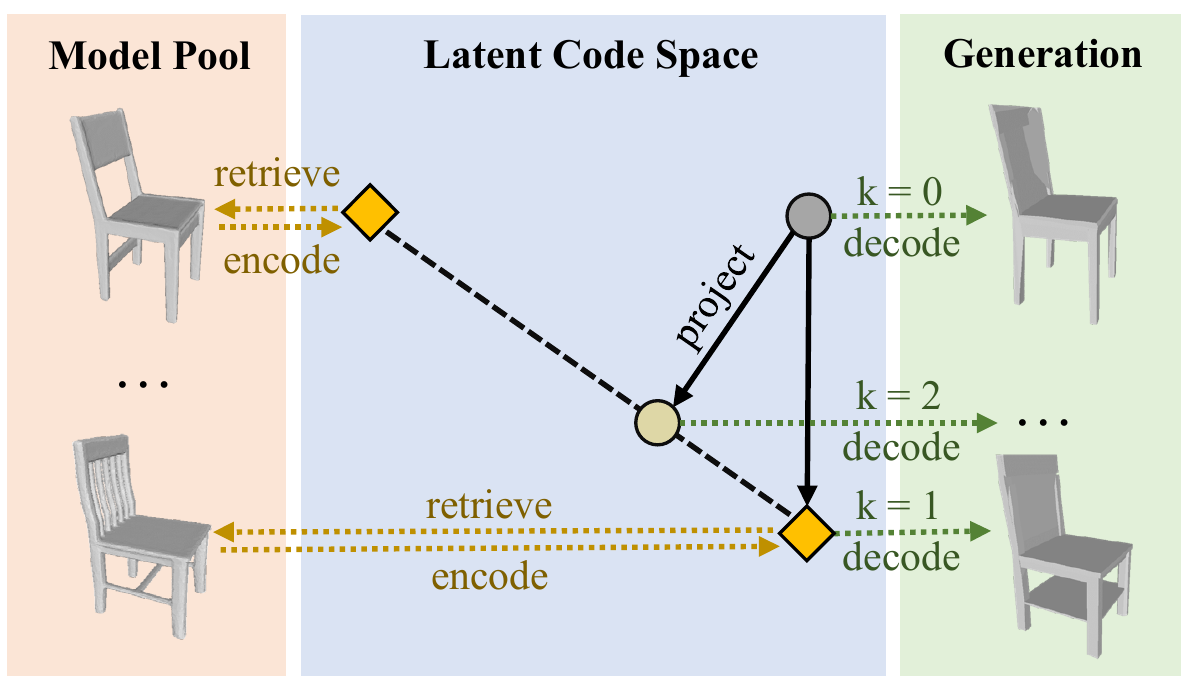}
        \label{fig:interp}
    }
    
    \caption{
    (a) Instead of explicitly predicting a complete mesh in 3D space, we leverage a mesh-aware latent code space to disentangle shape completion and mesh generation.
    (b) The gray circle is the predicted latent code from the observed partial point cloud. 
    We can directly use it to generate a mesh via the decoder (green right arrow with $k = 0$).
    With access to the CAD meshes or their latent codes (yellow diamonds) from the model pool, we can further perform model retrieval (yellow left arrows) or assisted generation by projecting the latent code to the nearest $k$ neighbors (green right arrows with $k > 0$).
    }
\end{figure}

The loss function for the second stage $\mathcal L^\text{stage2}$ further adds three regression terms on the basis of $\mathcal L^\text{stage1}$, \textit{i.e.}, the confidence loss $\mathcal L_\text{reg}^\text{conf}$, the bounding box loss $\mathcal L_\text{reg}^\text{bbox}$, and the latent distribution loss $\mathcal L_\text{reg}^\text{latent}$, all using the weighted smooth L1 loss to alleviate the class imbalance in object proposals.

\subsection{Disentangled Instance Mesh Reconstruction}
\label{sec:meshgen}
Instance mesh reconstruction from a partially observed point cloud is a challenging task, requiring both shape completion and mesh generation.
Previous methods perform these two tasks as a whole, but fail to generate high fidelity meshes that match the observation.
The problems are two-fold:
1) The difficulty of optimizing a conditioned mesh generator with the detection network, where lots of false positive proposals are used as inputs.
2) The ambiguity of learning complete shapes from incomplete point cloud observations.
To handle these problems, we propose a disentangled mesh reconstruction approach as illustrated in Figure~\ref{fig:iimr}.
The core idea here is to disentangle shape completion and mesh generation into two stages.
First, we pre-train a mesh Variational Autoencoder (VAE) to encode 3D meshes into a latent code space, which can be viewed as a mesh generator.
Note that this mesh generator is trained with complete GT meshes, and there is no ambiguity in shape learning since no completion happens here.
For shape completion, we can simply train another encoder that maps incomplete instance proposals into the same latent code space.
This completion encoder is supervised with these low-dimension latent codes as described in Section~\ref{sec:proposalwise}, which are much easier to optimize compared to high-dimension conditioned occupancy values.
Thus, the mesh generation part is detached from the training process of object recognition and shape completion.

We adopt BSP-Net~\cite{chen2020bspnet,bspnet_tpami}, which proposes an efficient approach to approximate low-poly meshes by learning convex decomposition, as the autoencoder model.
Specifically, we adopt a Conditional VAE (CVAE) variant for better generation quality.
Following~\cite{chen2020bspnet}, we sample complete point clouds from the CAD models and voxelize them as the input to the encoder, 
which outputs a latent distribution $\mathcal N(\boldsymbol \mu, \boldsymbol \sigma)$ that characterizes the shape.
Then, we sample a latent code $\mathbf z$ from this distribution,
and use the decoder to output a set of planes with the convex decomposition to generate the polygonal meshes.
In addition to the original BSP loss $\mathcal L^\text{bsp}$ in~\cite{chen2020bspnet}, a KL loss $\mathcal L^{\text{KL}}$ weighted by $0.1$ is added as regularization. 
More details can be found in the supplementary material.

After convergence, we use the encoder to extract the ground-truth latent distribution for each ground-truth mesh $M^\text{gt}_j$:
\begin{equation}
    \boldsymbol \mu^\text{gt}_j, \boldsymbol \sigma^\text{gt}_j = \text{Enc}(M^\text{gt}_j)
\end{equation}
Therefore, to reconstruct mesh $M_j$ from a partially observed point instance, we only need to regress the latent distribution and sample a latent code $\mathbf z_j \sim \mathcal N(\boldsymbol \mu_j, \boldsymbol \sigma_j)$, then decode it through: 
\begin{equation}
    M_j = \text{Dec}(\mathbf z_j)
\end{equation}
Optional post-processing like the Iterative Closest Point (ICP) algorithm can be used to further fine-tune the mesh location.
By default, we use the expectation as the latent code so $\mathbf z_j = \boldsymbol \mu_j$, 
but we can also sample different latent codes for different explanations of the partial point observation to address the ambiguity problem.
Furthermore, if we have access to the model pool $\mathcal M$ at test time, 
our model can also perform CAD retrieval task without any further training, by searching the nearest mesh from the model pool in the latent space: 
\begin{equation}
    M^\text{retr}_j = \arg \min_{m \in \mathcal M} ||\mathbf z_m - \mathbf z_j||_2
\end{equation}
where $\mathbf z_m$ is the latent code of CAD model $m$.
However, maintaining the whole model pool requires extra storage (about 238MB for the 2238 models used in Scan2CAD).
Another option is to only maintain the latent codes $\{\mathbf z_m | m \in \mathcal M\}$, which is a $256$-d vector for each mesh (about 2.2MB for the same models).
These latent codes can serve as priors to assist the mesh generation, by projecting the predicted latent code to the hyperplane spanned by the nearest $k$ latent codes $\{\mathbf z_{n1}, \mathbf z_{n2}, \cdots, \mathbf z_{nk} \}$ from the model pool:
\begin{equation}
    M^\text{proj}_j = \text{Dec}(\text{proj}_{\text{span}\{\mathbf z_{n1}, \mathbf z_{n2}, \cdots, \mathbf z_{nk} \}}(\mathbf z_j))
\end{equation}
Figure~\ref{fig:interp} illustrates the relationship between these methods.

\section{Experiment}
\begin{figure*}[ht!]
    \centering
    \includegraphics[width=\textwidth]{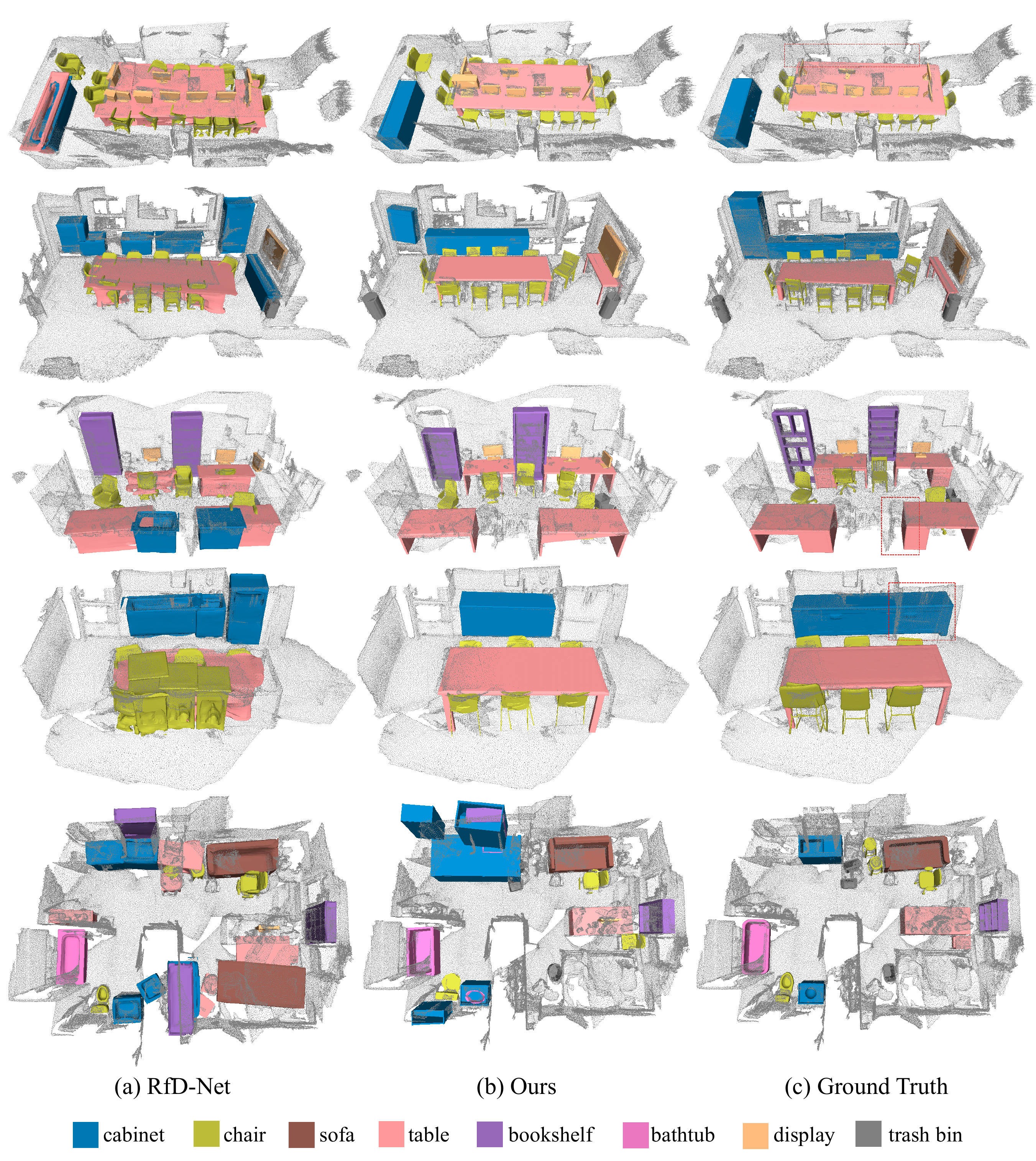}
    \caption{{Qualitative Comparison on the ScanNet dataset.} 
    Please zoom in to see details.
    We only visualize meshes with confidence larger than 0.5 for RfD-Net as in~\cite{Nie_2021_CVPR}, and 0.3 for Ours.
    Red-dash boxes show missing or incorrect human annotations in the ground truth, \textit{e.g.}, missing chairs, incorrectly scaled table and cabinet.
    More visualizations can be found in the supplementary material.}
    \label{fig:quality}
\end{figure*}

\begin{figure*}[ht!]
    \centering
    \includegraphics[width=\textwidth]{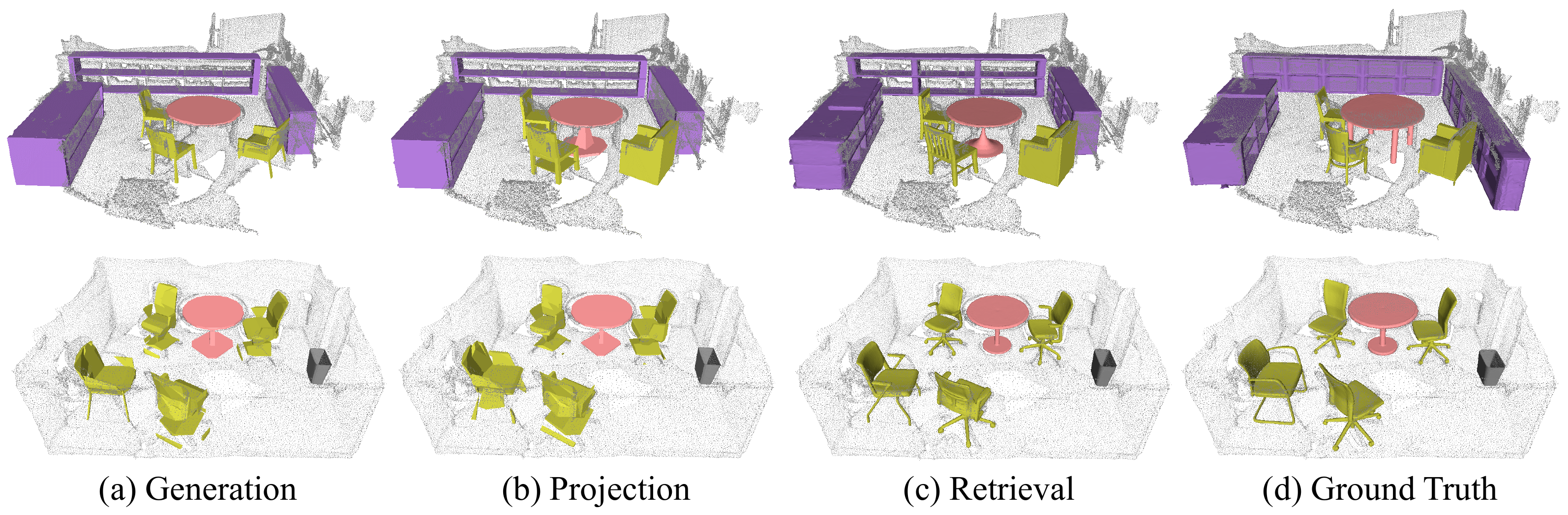}
    \caption{{Comparison between mesh generation, assisted generation (projection) and retrieval mode of our method.} }
    \label{fig:retrieve}
\end{figure*}

\subsection{Experimental Settings}

\noindent \textbf{Datasets.} 
Three datasets are involved in our experiments: ScanNet v2~\cite{dai2017scannet}, ShapeNet~\cite{chang2015shapenet}, and Scan2CAD~\cite{avetisyan2019scan2cad}.
The ScanNet dataset consists of 1,513 real world indoor scene scans. 
Inline with~\cite{Nie_2021_CVPR,hou2020revealnet}, The official data split is used in all experiments.
Only the incomplete point clouds are used as the input data, with the semantic and instance labels as the point-level supervision.
The Scan2CAD dataset provides the alignment of CAD models in ShapeNet to scenes in ScanNet.
We use the aligned CAD models as the proposal-level supervision.

It's worth noting that the semantic and instance labels of ScanNet point clouds are inconsistent with Scan2CAD meshes. 
The point cloud instance segmentation literature~\cite{jiang2020pointgroup,he2021dyco3d,chen2021hierarchical} usually adopts a 20-class label system, with 2 stuff categories and 18 object categories.
However, the instance mesh reconstruction literature~\cite{Nie_2021_CVPR} only uses 8 object categories.
To bridge between these two tasks, we make a compatible label system to both tasks and relabel the instance segmentation ground truths based on the new label system.
Note that we don't introduce any new manual labeling, but simply rearrange the two sources of label information.
The details of the label mapping and relabeling can be found in the supplementary material.

\noindent \textbf{Evaluation Metrics.}
The quality of point scene instance mesh reconstruction could be evaluated in two aspects: the completion quality and the mapping quality. The completion quality measures how well the reconstructed meshes match the human-annotated ground-truth meshes, and the mapping quality measures how well the reconstructed mesh surfaces match the observed point clouds.

\textit{Completion quality.}
Several approaches have been proposed to measure the similarity between different meshes. 
For voxel-based methods, \cite{Nie_2021_CVPR} voxelizes the meshes with a fixed voxel resolution and calculates the 3D Intersection over Union (IoU) of the prediction and the ground truth.
For point-based methods, \cite{chen2019learning,chen2020bspnet} sample point clouds with a fixed number from the mesh surfaces and calculate the Chamfer Distance (CD).
For mesh-based methods, \cite{chen2020bspnet,chen2003visual} choose to render multi-view 2D images and calculate the Light Field Distance (LFD).
All of these metrics can only partially represent the underlying mesh similarity.
For example, 3D IoU requires the predicted mesh to occupy the exact voxels as the ground-truth mesh, 
CD measures the distance between mesh surfaces, and LFD compares more on visual appearance.
Previous works~\cite{Nie_2021_CVPR,hou2020revealnet} only adopt 3D IoU as the metric, but we argue that 3D IoU at a coarse voxel resolution fails to reflect the quality of mesh surface and visual appearance, which are also important for measuring mesh similarity.
For a thorough comparison, we adopt all the three metrics with different thresholds to determine whether a predicted mesh can match a ground-truth mesh, and report the 3D detection mean AP over all classes.

\textit{Mapping quality.} As shown in Figure~\ref{fig:teaser}, the human-annotated ground-truth meshes may not be the only plausible reconstruction due to the ambiguity of input. Therefore, only using the completion quality may lead to biased evaluation. 
To alleviate this problem, we use mapping quality to measure how well the reconstructed mesh surfaces match the observed point clouds.
Specifically, we propose the Point Coverage Ratio (PCR), which computes the nearest distance from each observed instance point to the corresponding mesh surface,
and uses a threshold to determine whether this point belongs to the reconstructed surface:
\begin{equation}
    \text{PCR} = \frac {1} {|\mathcal P|} \sum_{\mathbf p \in \mathcal P} \mathds{1}_{\{ \text{dist}(\mathbf p, \mathcal M) < \omega \}}
\end{equation}
where $\mathbf p$ is a point from the observed ground-truth instance point cloud $\mathcal P$, 
$\mathds{1}$ means the indicator function,
$\mathcal M$ is the reconstructed mesh,
$\text{dist}(\mathbf p, \mathcal M)$ is the Euclidean distance from the point to the mesh surface,
and $\omega$ is the distance threshold.
A larger PCR means more observed points are located near the mesh surface, and thus better mapping quality.
Similarly, we report the 3D detection mean AP over all classes using PCR as the metric.

\noindent \textbf{Baselines.}
We mainly compare our results with the current state-of-the-art RfD-Net~\cite{Nie_2021_CVPR}, which is the first work that generates high-resolution mesh for point scene instance reconstruction.
The officially released model is used to generate results for evaluation and comparison.
Also, we investigate different variants of our method including direct mesh generation, assisted mesh generation by latent code projection and mesh retrieval with different levels of access to the CAD model pool. 
All methods in our experiments are trained on the same dataset split, and evaluated with the same hyper-parameters.

\noindent \textbf{Implementation Details.}
We set the voxel size as $0.02$m for point-wise sparse U-Net, and $0.05$m for proposal-wise sparse U-Net.
We use $D_\text{point} = 32$, $D_\text{prop} = 64$, and $D_\text{shape} = 128$.
We train 256 epochs for the first stage, and 256 epochs for the second stage with a batch size of 8 on a single Nvidia Tesla V100 GPU.
The Adam optimizer is used with an initial learning rate of $0.001$ for the first stage and $0.0001$ for the second stage.
In training, we set the clustering radius as 0.03m following PointGroup~\cite{jiang2020pointgroup}.
The weight for each loss term defaults to $1.0$.
In evaluation, we use a multi-scale clustering method at $\{0.01, 0.03, 0.05\}$m to spot more proposals.
The nearest neighbour count $k$ is set to $1$ for the projection model.
The voxel size for 3D IoU calculation and the threshold $\omega$ for PCR calculation are both set to $0.047$m following~\cite{Nie_2021_CVPR}.
The proposal confidence threshold is set to $0.09$, and each proposal should have at least $100$ points.

\begin{table*}[t!]
    \small
    \caption{\textbf{Comparisons on mesh completion quality.} We report mean AP for different metric@threshold. For IoU, higher threshold is more difficult. For CD and LFD, smaller thresholds are more difficult.
    Better results are in bold. We don't compare with projection and retrieval models since they use extra information.
    }
    \label{tab:completion}    
    \begin{center}
        \setlength{\tabcolsep}{4pt}
        \begin{tabular}{l | cccccc }
            \shline
                                         & IoU@0.25  & IoU@0.5   & CD@0.1  & CD@0.047    & LFD@5000  & LFD@2500 \\
            \hline
            RfD-Net~\cite{Nie_2021_CVPR} & 42.52     & \bf 14.35 & 46.37     & 19.09     & 28.59     & 7.80     \\
            \hline
            Ours                      & \bf 46.34 & 12.54     & \bf 52.39 & \bf 25.71 & \bf 29.47 & \bf 8.55     \\
            \hline
            Ours $+$ proj.                    & 46.50     & 12.59     & 52.06     & 24.84     & 29.95     & 9.67     \\ 
            Ours $+$ retr.                    & 47.20     & 12.83     & 51.77     & 25.10     & 30.80     & 10.12    \\ 
            \shline
        \end{tabular}
    \end{center}
    \vspace{-0.5cm}
\end{table*}

\begin{table*}[t!]
    \small
	\caption{\textbf{Comparisons on point-to-mesh mapping quality.}
	The AP scores are measured with PCR@0.5.
	}
	\label{tab:mapping}
	\begin{center}
		\begin{tabular}{l | c c c c c c c c | c}
			\shline
			& table & chair & bookshelf & sofa & trash bin & cabinet & display & bathtub & mean \\
			\hline
			Scan2CAD~\cite{avetisyan2019scan2cad} & 36.60     &     69.31 & \bf 65.03 &     28.92 &     56.93 & \bf 41.82 &     70.81 &     45.07 &     45.07 \\
			RfD-Net~\cite{Nie_2021_CVPR}    & 32.54     &     76.54 &     30.66 &     22.91 &     40.54 &     24.37 &     67.64 &     52.69 &     43.49 \\
			\hline
			Ours                         & \bf 49.78 & \bf 78.64 &     29.25 & \bf 60.33 & \bf 65.30 &     18.75 & \bf 76.56 & \bf 75.51 & \bf 56.76 \\
			\hline
			Ours $+$ proj.                       & 60.57     &     78.88 &     28.93 &     61.00 &     65.61 &     18.45 &     78.02 &     72.79 &     58.03 \\
			Ours $+$ retr.                       & 62.28     &     75.57 &     45.23 &     52.60 &     65.27 &     17.14 &     76.61 &     73.81 &     58.82 \\
			\shline
		\end{tabular}
	\end{center}
	\vspace{-0.5cm}
\end{table*}
\begin{table*}[t!]
    \small
    \caption{\textbf{Object Recognition Precision.} 
    We report the precision of object recognition at different IoU thresholds. Our method greatly reduces the number of false positive proposals.
    }
    \label{tab:fp}    
    \begin{center}
        \setlength{\tabcolsep}{4pt}
        \begin{tabular}{l | cc }
            \shline
                                         & Prec.@0.25 & Prec.@0.5 \\
            \hline
            RfD-Net~\cite{Nie_2021_CVPR} &  22.99 & 7.92 \\ 
            \hline
            Ours                      &  \textbf{43.70} & \textbf{15.09} \\
            \shline
        \end{tabular}
    \end{center}
    \vspace{-0.5cm}
\end{table*}

\subsection{Comparisons}

\noindent \textbf{Quantitative Comparisons.}
Table~\ref{tab:completion} shows the quantitative comparisons of completion quality. 
Our method outperforms state-of-the-art on five out of six evaluation settings in terms of mAP. 
This benefits from the proposed pipeline that reduces false positive proposals (Table~\ref{tab:fp}) and enhances the quality of generated meshes through the disentangled mesh reconstruction approach.
A reason for the lower 3D IoU with $0.5$ as the threshold might be that 3D IoU with a high threshold discourages meshes with thin structures, since a small displacement can result in huge drop in the metric value, even if the shape is of good quality (\eg the chair in Figure~\ref{fig:teaser}).
With the access of an external model pool, the projection and retrieval models (denoted as Ours $+$ proj. and Ours $+$ retr.) produce more robust meshes.
In particular, the retrieval model achieves a better LFD score, since the retrieved meshes are guaranteed to be rational.

Table~\ref{tab:mapping} shows the quantitative comparisons of mapping quality. 
We achieve significant improvement in mAP and surpass the human-annotated Scan2CAD dataset. 
This is to be expected. 
Since the ShapeNet dataset is synthetic and has a finite number of models, the human-annotated CAD models may not perfectly match the point cloud observations.
Interestingly, the results also partially reveal the capability of the CAD model pool.
For example, retrieved meshes show better performance on bookshelf, but perform worse on sofa, 
which means there may be fewer suitable sofa CAD models that fit real world ScanNet data. 
In such cases, generated meshes can be potentially better.

\noindent \textbf{Qualitative Comparisons.}
Figure~\ref{fig:quality} shows the qualitative comparisons.
The meshes generated by our method have better visual appearance and more accurate locations.
Besides, we show that when the human-annotated ground truths conflict with the observed point clouds, our model can still successfully detect these instances and output plausible meshes.
In Figure~\ref{fig:retrieve}, we also show the results of the projection and retrieval models.
With extra information from the model pool, the model can produce more robust meshes.

\subsection{Ablation Study}

We conducted ablation studies to verify the influence of proposed modules in Table~\ref{tab:ablation}.
`ResBox' means we learn a residual bounding box $\Delta \mathbf b_j$ to refine the empirical bounding box deduced from observed point clouds.
`MSC' means we use multiple clustering radii to find more proposals at test time.
`ICP' means we apply the ICP algorithm to post-process the reconstructed meshes.
The results indicate that the combination of these three modules achieves overall the best performance.
In particular, the proposed residual bounding box learning refines the object location and improves all metrics, while ICP post-processing mainly affects the IoU metric.

\begin{table}[t]

\caption{\textbf{Ablation study of proposed modules.}}
\label{tab:ablation}

\begin{center}
\setlength{\tabcolsep}{4pt}
\begin{tabular}{c c c c c c}
	\shline
	ResBox & MSC    & ICP    & IoU@0.25  & CD@0.1    & LFD@5000\\
	\hline
	       & \cmark & \cmark &     41.76 &     46.51 & 27.55 \\
	\cmark &        & \cmark &     43.57 &     48.26 & 28.91 \\
	\cmark & \cmark &        &     42.33 &     52.19 & 29.11 \\
    \cmark & \cmark & \cmark & \bf 46.34 & \bf 52.39 & \bf 29.47\\ 
	\shline
\end{tabular}
\end{center}
\vspace{-0.5cm}

\end{table}

\section{Conclusion}
In this paper, we introduce a Disentangled Instance Mesh Reconstruction pipeline for point scene understanding. 
Our method first performs instance segmentation to generate accurate object proposals, then applies a disentangled instance mesh reconstruction strategy to mitigate the ambiguity of learning complete shapes from incomplete point observations.
We evaluate the experimental results on the challenging ScanNet dataset from the perspectives of completion quality and mapping quality, and demonstrate the superior performance of our method.

\noindent \textbf{Acknowledgements.} 
This work is supported by the National Key Research and Development Program of China (2020YFB1708002), National Natural Science Foundation of China (61632003, 61375022, 61403005), Beijing Advanced Innovation Center for Intelligent Robots and Systems (2018IRS11), and PEK-SenseTime Joint Laboratory of Machine Vision.

\clearpage

\bibliographystyle{splncs04}
\bibliography{egbib}

\begin{appendix}
\appendix

\section{Data Processing}
In this section, we detail the data processing in our experiments.

\subsection{Semantic Segmentation Labels}
\begin{figure}[t!]
    \centering
    \includegraphics[width=\textwidth]{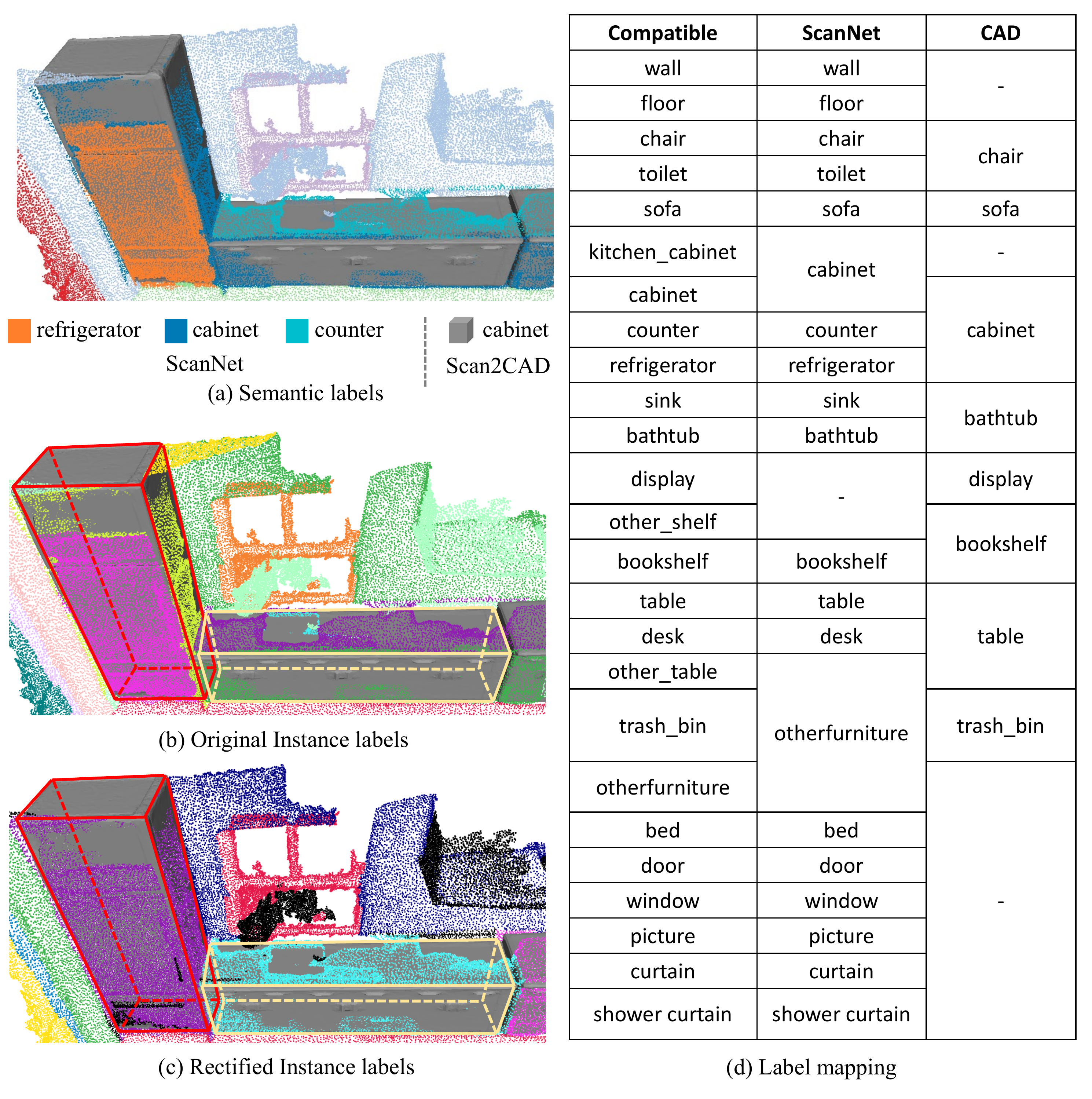}
    \caption{\textbf{Semantic and Instance Segmentation Labels.} 
    (a) The \textit{cabinet} class in Scan2CAD may refer to \textit{cabinet, counter, refrigerator} in ScanNet.
    (b) The original instance labels in ScanNet. A single object is divided into multiple parts, such as the refrigerator marked with the red box.
    (c) The rectified instance labels are consistent with Scan2CAD meshes. 
    (d) Label mapping between different tasks.
    }
    \label{fig:class_match}
\end{figure}

To enable training with both ScanNet point-level annotations and Scan2CAD instance-level annotations, we first make a compatible label system that contains 25 semantic classes, including 2 stuff classes and 23 object classes. 
This label system basically extends the ScanNet 20-class label system by subdividing some classes, to make sure we can convert safely to the CAD 8-class label system.
The full label mapping is shown in Figure~\ref{fig:class_match} (d).
For the semantic segmentation task, we use all the 25 classes as the supervision.

\subsection{Instance Segmentation Labels}
Due to the complicated containment relationships between these two label systems as shown in Figure~\ref{fig:class_match} (d), 
the ScanNet instance segmentation labels are still inconsistent with Scan2CAD models.
For example, the \textit{cabinet} class in Scan2CAD may refer to \textit{cabinet, counter, refrigerator} in ScanNet instance segmentation annotations.
Directly using the ScanNet annotations may lead to inaccurate instance supervision for our task, such as the instance centers and point-wise offsets.
Figure~\ref{fig:class_match} (a) and (b) shows an example of such a condition.
Therefore, we need to merge the instances that belong to the same CAD class.
Since these instances mainly belong to the \textit{cabinet} class, we use the instance bounding boxes from the Scan2CAD dataset to rectify the ScanNet instance segmentation annotations, as illustrated in Figure~\ref{fig:class_match} (c).
These new instance labels are used to supervise the instance segmentation task in our pipeline. 
Please note that we do not manually annotate this dataset, but only rearrange the existing annotations to make the training consistent.

\section{Experimental Results}

Besides the results analyzed in the main paper, we conduct more ablations on the proposed method and show more qualitative comparisons.

\subsection{More Ablations on the Proposed Method}

\begin{figure*}[h]
    \centering
    \includegraphics[width=\linewidth]{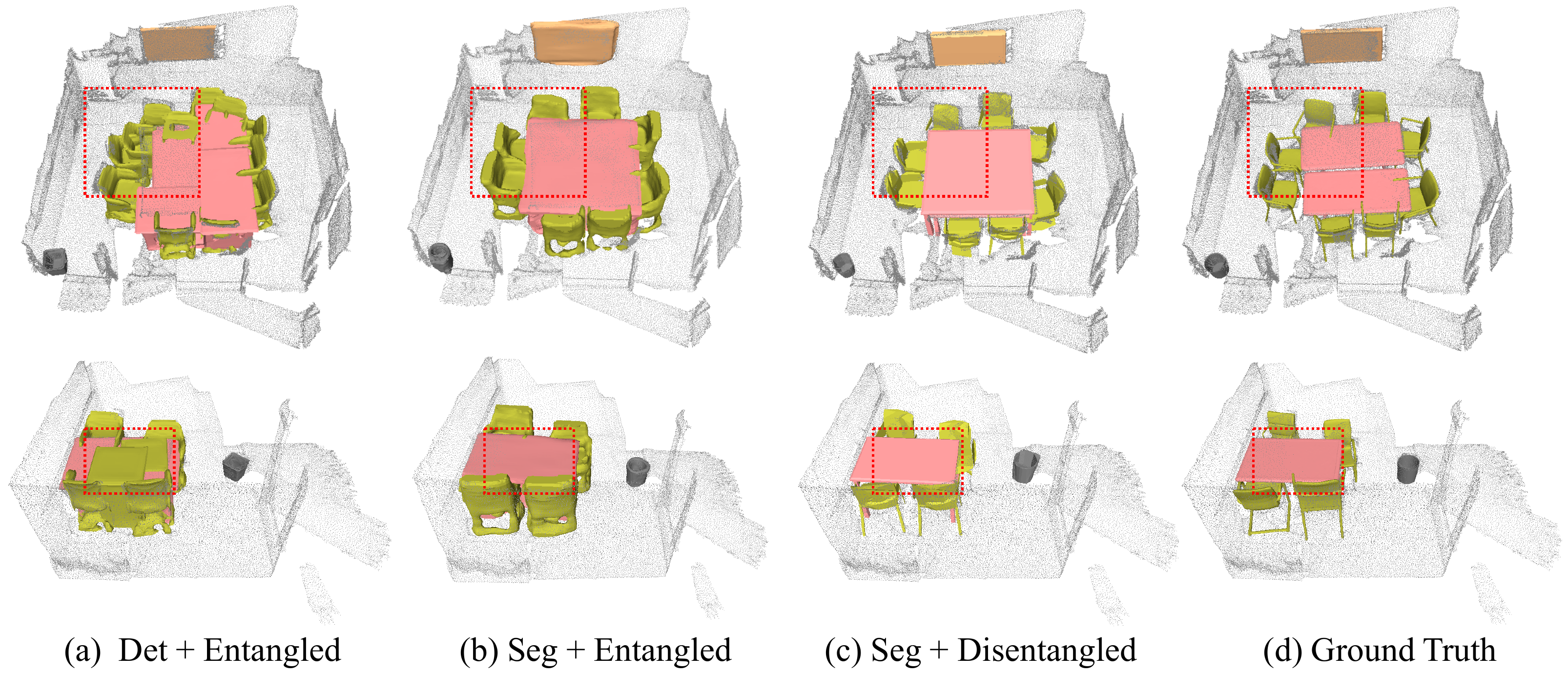}
    \caption{{More qualitative comparisons.}
    The segmentation backbone can improve the quality of proposals by reducing false positives as marked by the red boxes, and the disentangled mesh reconstruction strategy improves the quality of generated meshes.
    }
    \label{fig:more}
\end{figure*}

\begin{table}[t]

\caption{\textbf{Ablation study on the proposed method.} 
RfD-Net is referred to as `Det + Entangled', and our method is referred to as `Seg + Disentangled'.
}
\label{tab:more_ablation}

\begin{center}
\setlength{\tabcolsep}{4pt}
\begin{tabular}{l | c c c}
	\shline
	pipeline           & IoU@0.25  & CD@0.1    &    LFD@5000\\
	\hline
	Det + Entangled    &     42.52 &     46.37 &    28.59 \\
	Seg + Entangled    &     43.95 &     47.46 &    28.67 \\
	Seg + Disentangled &     \textbf{46.34} &     \textbf{52.39} &    \textbf{29.47} \\
	\shline
\end{tabular}
\end{center}

\end{table}

We perform ablation studies to verify the two contributions of the proposed method: 
(i) instance segmentation $\rightarrow$ instance mesh completion pipeline,
and (ii) disentangled instance mesh reconstruction (DIMR) strategy.
Results are listed in Table~\ref{tab:more_ablation}.
We design three experiments: 
(i) `Det + Entangled',
(ii) `Seg + Entangled',
 and (iii) `Seg + Disentangled'.
`Entangled' means directly learning occupancy values of complete objects based on incomplete point observations, which is proposed in \cite{Nie_2021_CVPR}. `Disentangled' means our DIMR strategy.
RfD-Net~\cite{Nie_2021_CVPR} is referred to as `Det + Entangled' and the proposed method is referred to as `Seg + Disentangled'.
Comparing first two rows in Table~\ref{tab:more_ablation},
we find that the segmentation-then-completion pipeline achieves slightly better performance than detection-then-completion pipeline, due to the reduction of false positive proposals (we also show some visualizations in Figure~\ref{fig:more} to verify this).
Furthermore, when we adopt the disentangled instance mesh reconstruction strategy, the performance improves significantly, $e.g.$, from 43.95 to 46.34 in terms of IoU@0.25. Overall, DIMR makes the largest contribution to the performance gain.

\subsection{More Visual Comparisons between the Variants of Our Method}
\begin{figure*}[t]
    \centering
    \includegraphics[width=\linewidth]{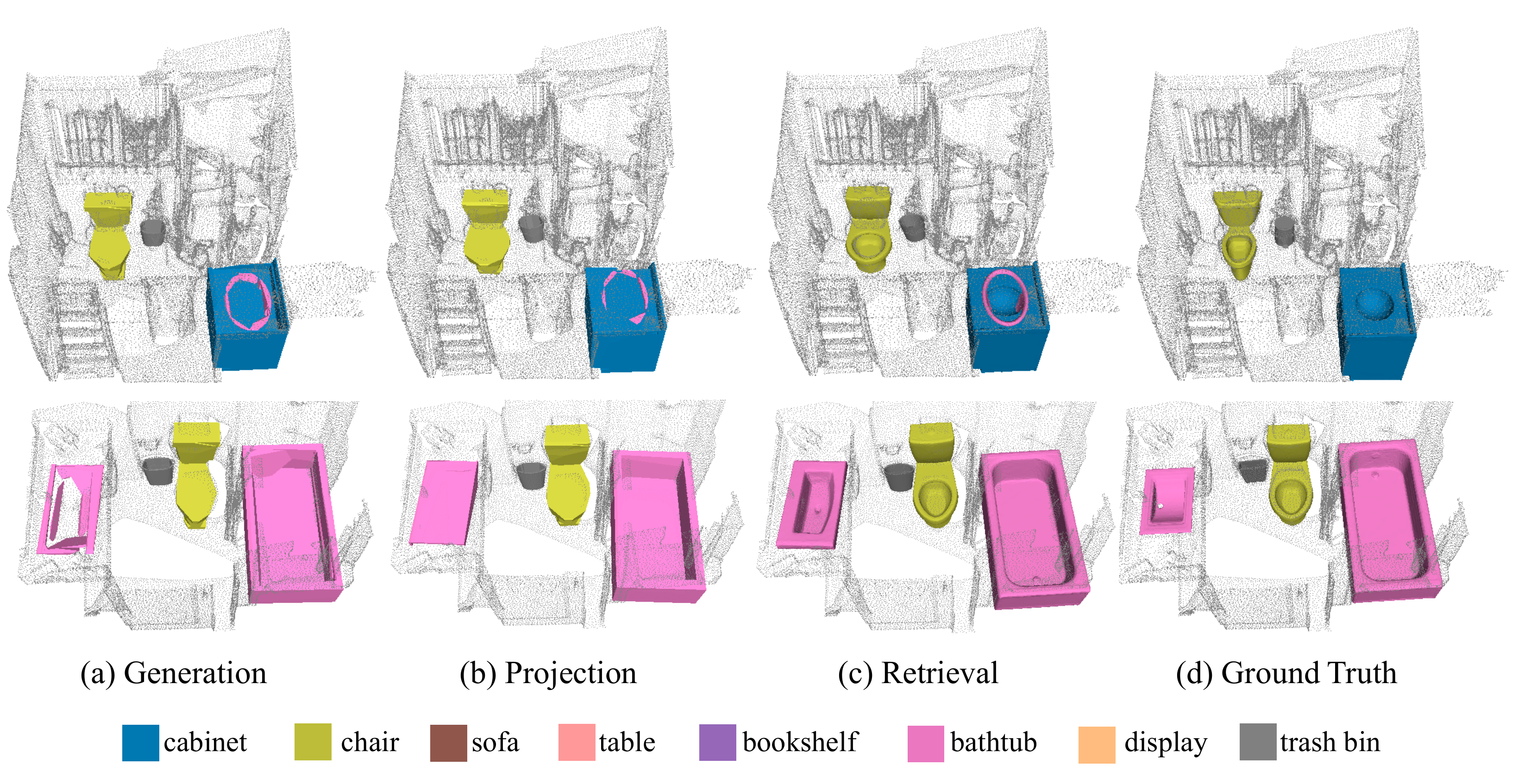}
    \caption{{More comparisons on mesh generation, mesh retrieval and assisted mesh generation.}
    }
    \label{fig:more_ret}
\end{figure*}

We also provide more comparisons between the three variants of our method in Figure~\ref{fig:more_ret}. 
The BSP mesh decoder used in our generation pipeline suffers from generating concave objects such as the toilet, but our retrieval variant can handle these objects well.

\begin{figure}[t]
    \centering
    \includegraphics[width=0.6\textwidth]{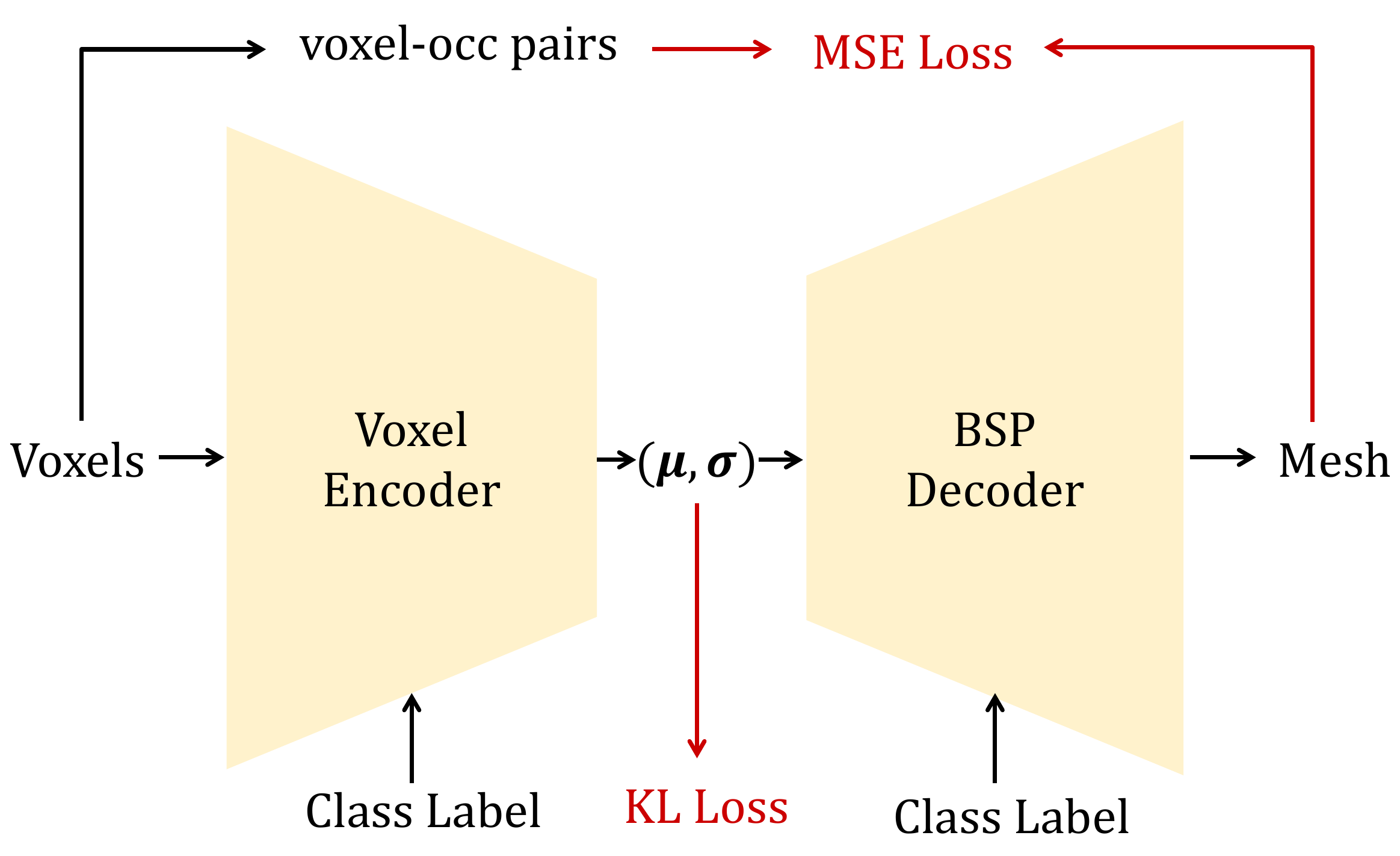}
    \caption{{BSP-CVAE architecture}.}
    \label{fig:bspcvae}
\end{figure}

\section{Details of the Mesh Autoencoder}
In this section, we discuss the details of the pre-trained mesh autoencoder, \textit{i.e.}, the BSP-CVAE network.

\begin{figure*}[t]
    \centering
    \includegraphics[width=\linewidth]{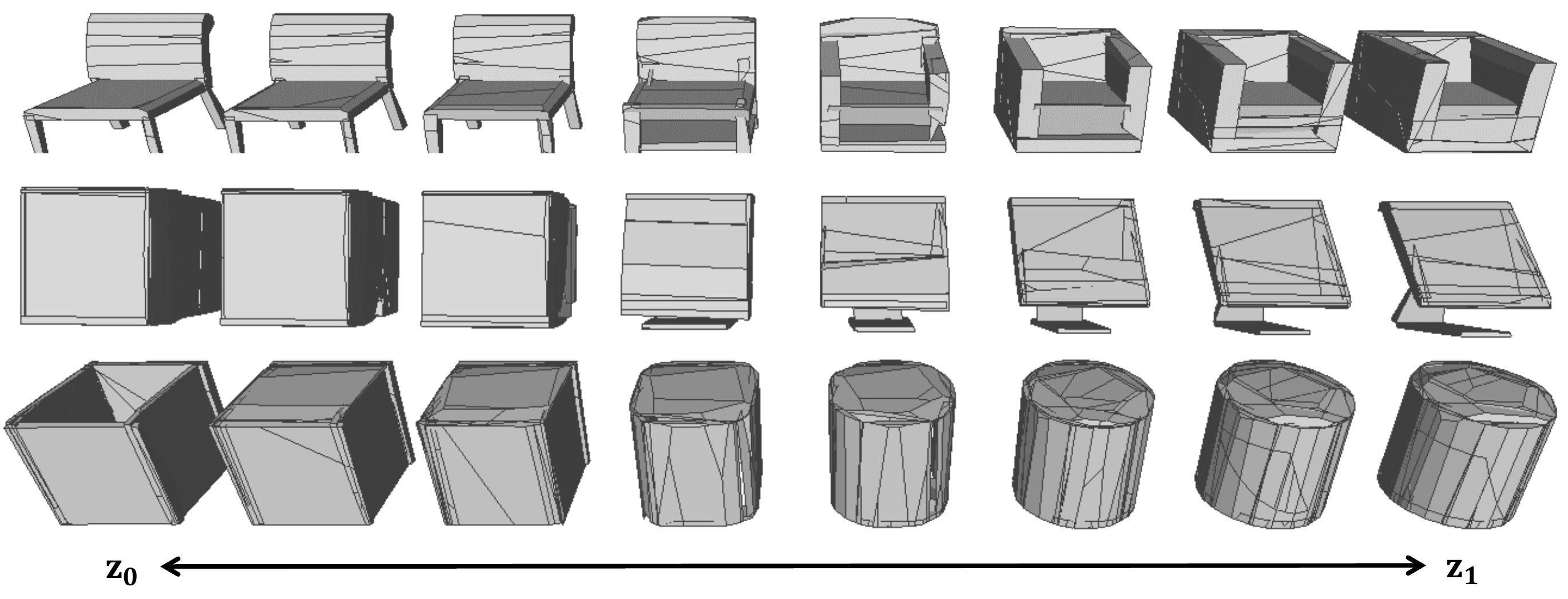}
    \caption{{Latent space interpolation results of BSP-CVAE.}
    We randomly choose two latent codes from ShapeNet meshes (left- and right-most) and interpolate linearly between them and visualize the generated meshes.
    }
    \label{fig:interp_bsp}
\end{figure*}

\subsection{Network Structure}
We follow~\cite{chen2020bspnet} to build a BSP-Net with some modifications, as illustrated in Figure~\ref{fig:bspcvae}.
The modifications we made include:
1) change the original autoencoder (AE) to a variational autoencoder (VAE), so the encoder outputs latent distributions and is supervised by an extra KL loss;
2) add the class label to the input, so the VAE turns to a conditional VAE (CVAE).
These make the network more suitable to our task, since we need to extract latent codes from class-aware CAD models and also generate 3D meshes given latent codes and class labels. 

\subsection{Training Details}
We use the CAD models from the ShapeNet dataset~\cite{chang2015shapenet} to train the network.
Specifically, we use the subset of eight classes provided by~\cite{Nie_2021_CVPR}.
For each CAD mesh, we voxelize it into $64 \times 64 \times 64$ volume as the input for the voxel encoder.
2048 empty voxels and 2048 occupied voxels are sampled from this volume at each training step to optimize the network.
We follow~\cite{chen2020bspnet} to train the BSP-CVAE for 800 epochs, including 400 epochs for the continuous phase and 400 epochs for the discrete phase.

\subsection{Latent Code Interpolation for BSP-CVAE}
To show the generative property of the trained BSP-CVAE, we provide some latent code interpolation experiments as shown in Figure~\ref{fig:interp_bsp}.
The model is capable of generating high quality meshes, and successfully learns the structural changes during interpolation of two given shapes, which meets our need for the mesh autoencoder.

\end{appendix}

\end{document}